\documentclass[twocolumn,10pt]{asme2e_hal}

\usepackage{graphicx}
\usepackage{caption}
\usepackage{subcaption}
\usepackage{amsmath}
\usepackage{amssymb}
\usepackage{mathrsfs}

\papernum{DETC2021/MR70233}

\title{A new robotic hand based on the design of fingers with spatial motions}
\author{Pol Hamon
    \affiliation{
    \begin{tabular}{cc}
	Armor M\'eca & École Centrale de Nantes/LS2N\\
    ZI La Grignardais  & UMR CNRS 6004, 1 rue de la Noe\\
    22490 Pleslin-Trigavou, France & 44321 Nantes, France\\
    \end{tabular}\\
     Email: Pol.Hamon@armor-meca.com} \\
   {\tensfb Damien Chablat,  Franck Plestan}     
    \affiliation{Ecole Centrale de Nantes, LS2N\\
    UMR CNRS 6004, 1 rue de la Noe, 44321 Nantes, France \\
    Email: Damien.chablat@cnrs.fr, Franck.Plestan@ec-nantes.fr}
}

\begin{document}

\maketitle  

\begin{abstract}
{\it This article presents a new hand architecture with three under-actuated fingers. Each finger performs spatial movements to achieve more complex and varied grasping than the existing planar-movement fingers. The purpose of this hand is to grasp complex-shaped workpieces as they leave the machining centres. Among the taxonomy of grips, cylindrical and spherical grips are often used to grasp heavy objects. A combination of these two modes makes it possible to capture most of the workpieces machined with 5-axis machines. However, the change in grasping mode requires the fingers to reconfigure themselves to perform spatial movements. This solution requires the addition of two or three actuators to change the position of the fingers and requires sensors to recognize the shape of the workpiece and determine the type of grasp to be used. This article proposes to extend the notion of under-actuated fingers to spatial movements. After a presentation of the kinematics of the fingers, the problem of stability is discussed as well as the transmission of forces in this mechanism. The complete approach for calculating the stability conditions is presented from the study of Jacobian force transmission matrices. CAD representations of the hand and its behavior in spherical and cylindrical grips are presented.}
\end{abstract}
\section{Introduction}
Robotic hands are inspired by human ones to enable robots to grasp parts or tools. They can be used for industrial robots, for humanoid robots or as prostheses for humans. 
Since the early 1980s, many hands have been designed as the Okada hand \cite{okada1982computer}, the Stanford/JPL hand \cite{salisbury1982articulated}, the Utah/MIT hand \cite{jacobsen1986design}, the LMS hand \cite{gazeau2001lms}. This first generation of hands suffered many disadvantages such as their cost and the complexity of their control. Other hands have been developed by reducing the number of actuators and the number of degrees of freedom \cite{akin2002development} \cite{crisman1996graspar}.
Another way is the creation of under-actuated hands allowing the number of actuators to be reduced without reducing the number of degrees of freedom \cite{laliberte1998simulation}.

The architecture of these under-actuated fingers has not changed since their origin and remains planar. In order to adapt to the objects to be grasped, a mechanism placed at the base of each finger must be used to switch from a cylindrical to a spherical or planar grasp \cite{laliberte1998simulation}. In this case, one or two actuators are required to change the type of grasp.

\begin{figure}
\centering
\includegraphics[width=6cm]{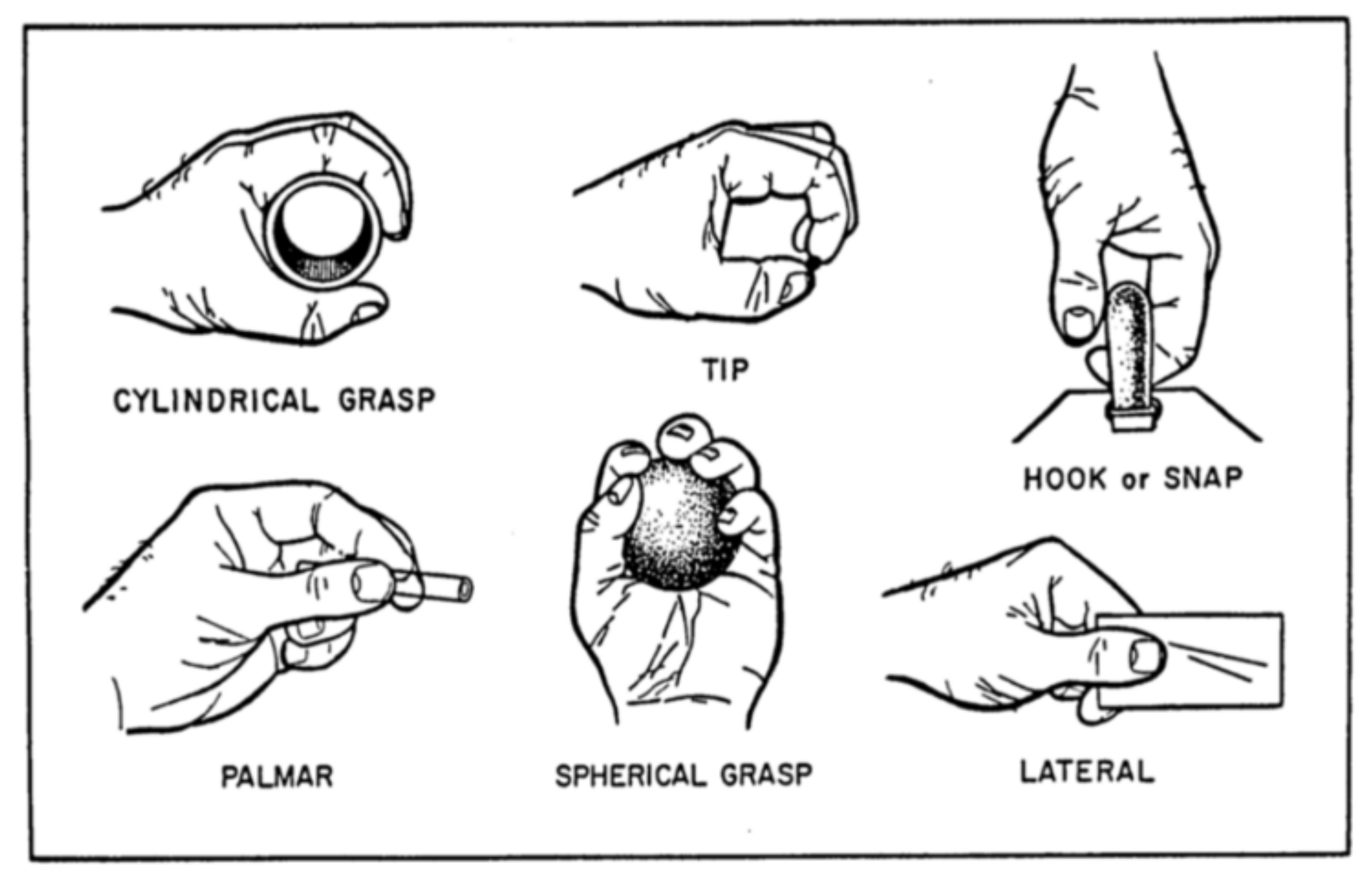}
\caption{Grasping of common objects of the daily life}
\label{fig:my_label}
\end{figure}

However, among the taxonomy of grips, the cylindrical and spherical grips are mainly used for gripping heavy objects \cite{cutkosky1989grasp}. The other types are for precision tasks and are not dedicated to the capture of heavy objects (Figure \ref{fig:my_label}). To simplify both design and control, the aim of this article is to introduce a new finger architecture that can naturally adapt to the shapes of objects.

Next section presents the architecture of the new fingers. The equations of the kinematics are presented after the description of the different closed chains. 
\section{Architecture of the new hand}
The purpose of this article is to present a new hand architecture, consisting of three fingers with a new architecture, the objective being to introduce a new type of spatial under-actuation. 
\subsection{Architecture of the new fingers}
\begin{figure}
    \center
    \includegraphics[width=7cm]{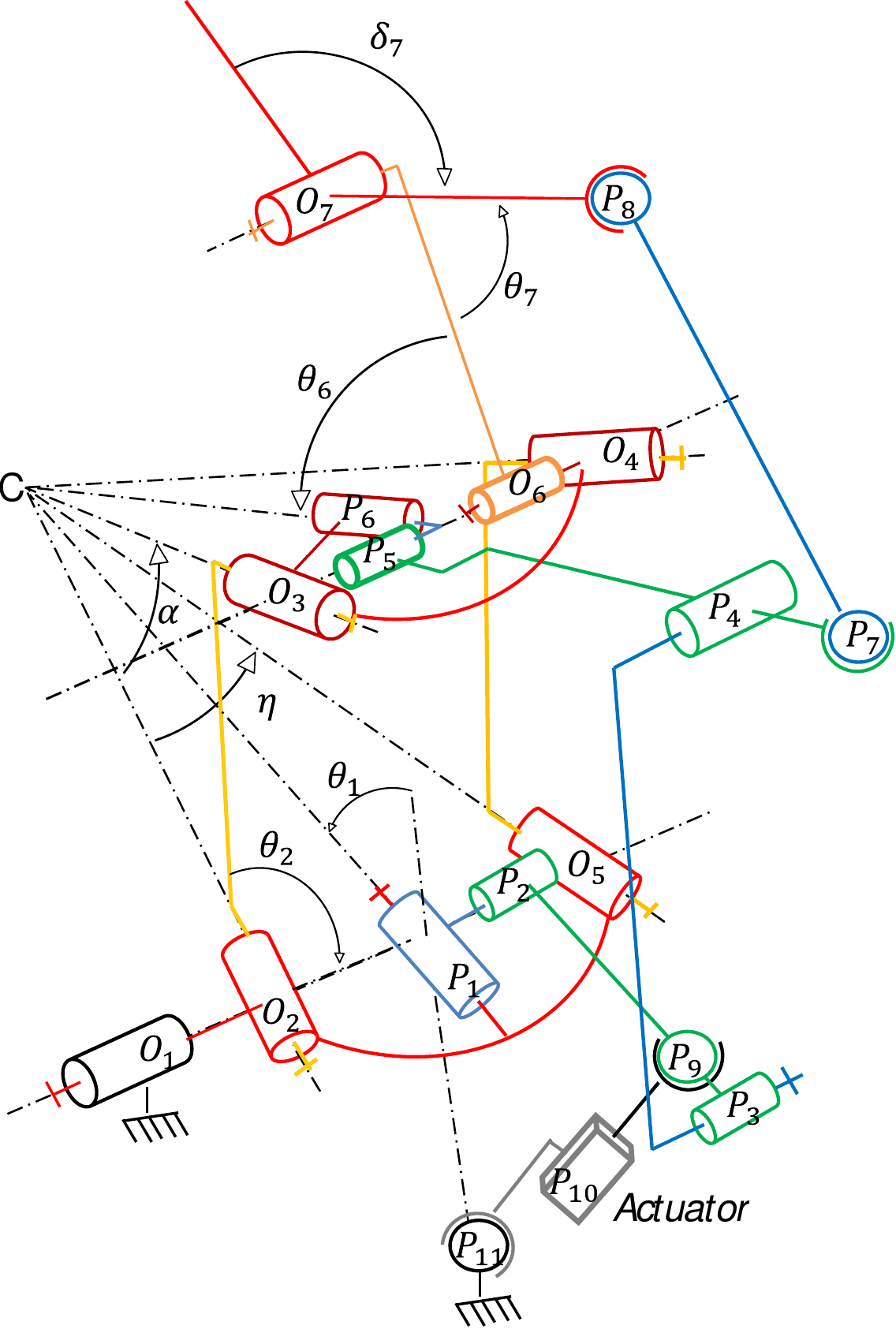}
    \caption{Kinematics of the new finger architecture}
    \label{fig:cinématique_de_la_nouvelle_architecture_de_main}
\end{figure}

To achieve spherical and cylindrical grasps, an under-actuated finger architecture
is presented featuring a spherical mechanism for the proximal phalanx \cite{mccarthy2010geometric} and four-bar mechanisms for the distal phalanges (Figure~\ref{fig:cinématique_de_la_nouvelle_architecture_de_main}). This solution provides a new degree of freedom from under-actuated fingers allowing a non-planar behaviour comparable to the abduction movements of the human hand and thus a change in the type of grip. The hand is an under-actuated three-fingers assembly whose placement is defined to achieve both types of grasp.

The innovation of this architecture comes mainly from the use of the spherical mechanism. 
In order to simplify the geometry of the spherical parallel mechanism, we set  
$||\overrightarrow{CO_{2}}||= ||\overrightarrow{CO_{3}}||= ||\overrightarrow{CO_{4}}||= ||\overrightarrow{CO_{5}}||$. 
In addition, the angles of the mechanism are two by two identical and defined as follows
$\widehat{O_{2}CO_{3}} = \widehat{O_{4}CO_{5}} = \alpha$ and $\widehat{O_{3}CO_{4}} = \widehat{O_{5}CO_{2}} = \eta$.

A new actuating system is adapted to this new kinematics in order to meet the stability requirements  when grasping objects. The four degrees of freedom of each finger can be defined as:
\begin{itemize}
    \item [$\bullet$] A first degree of freedom $\theta_{1}$ is defined as the movement of the proximal phalanx. $\theta_{1}$ is the angle between $\overrightarrow{P_{2}C}$ and $\overrightarrow{P_{2}P_{11}}$
    \item [$\bullet$] A second degree of freedom $\theta_{2}$ is defined as the movement of the spherical system.  $\theta_{2}$ is the angle between plans ($C,0_{2},0_{3}$) and ($C,0_{2},0_{5}$)
    \item [$\bullet$] A third degree of freedom $\theta_{6}$ is defined as the movement of the intermediate phalanx. $\theta_{6}$ is the angle between $\overrightarrow{0_{5} C}$ and $\overrightarrow{O_{6}O_{7}}$
    \item [$\bullet$] A fourth degree of freedom $\theta_{7}$ is defined as the movement of the distal phalanx. $\theta_{7}$ is the angle between $\overrightarrow{0_{7}0_{6}}$ and $\overrightarrow{O_{7}P_{8}}$
\end{itemize}
By using spherical trigonometry \cite{todhunter1863spherical} (Eq.~\eqref{theta1}) and the results coming from the study of parallel spherical mechanisms \cite{mccarthy2010geometric} (Eq.~\eqref{theta2}), the angles 
$\theta_1$, $\theta_2$, $\theta_6$ and $\theta_7$ read as:
\begin{eqnarray}
\theta_{1} &=& \frac{1}{2}\operatorname{acos}{\left(\frac{- \sin^{2}{\left(\frac{\eta}{2} \right)} + \cos{\left(\alpha \right)}}{\cos^{2}{\left(\frac{\eta}{2} \right)}}  \right)} \label{theta1}\\
\theta_{2} &=& - 2 \operatorname{atan}{\left(\sqrt{\frac{- \tan{\left(\frac{\alpha}{2} \right)} \tan{\left(\frac{\eta}{2} \right)} - 1}{\tan{\left(\frac{\alpha}{2} \right)} \tan{\left(\frac{\eta}{2} \right)} - 1}} \right)} \label{theta2}\\
\theta_{6} &=& - \frac{1}{2}\operatorname{acos}{\left(\frac{2 \cos{\left(\alpha \right)} + \cos{\left(\eta \right)} - 1}{\cos{\left(\eta \right)} + 1} \right)}-\pi/2 
\label{theta6}\\
\theta_{7} &=& \pi-\delta_{7} \label{theta7} 
\end{eqnarray}
The angles are defined as positive in the closing direction of the fingers.
\subsection{Grasping and fingers' positions}
\begin{figure}
\begin{center}
\includegraphics[width=5cm]{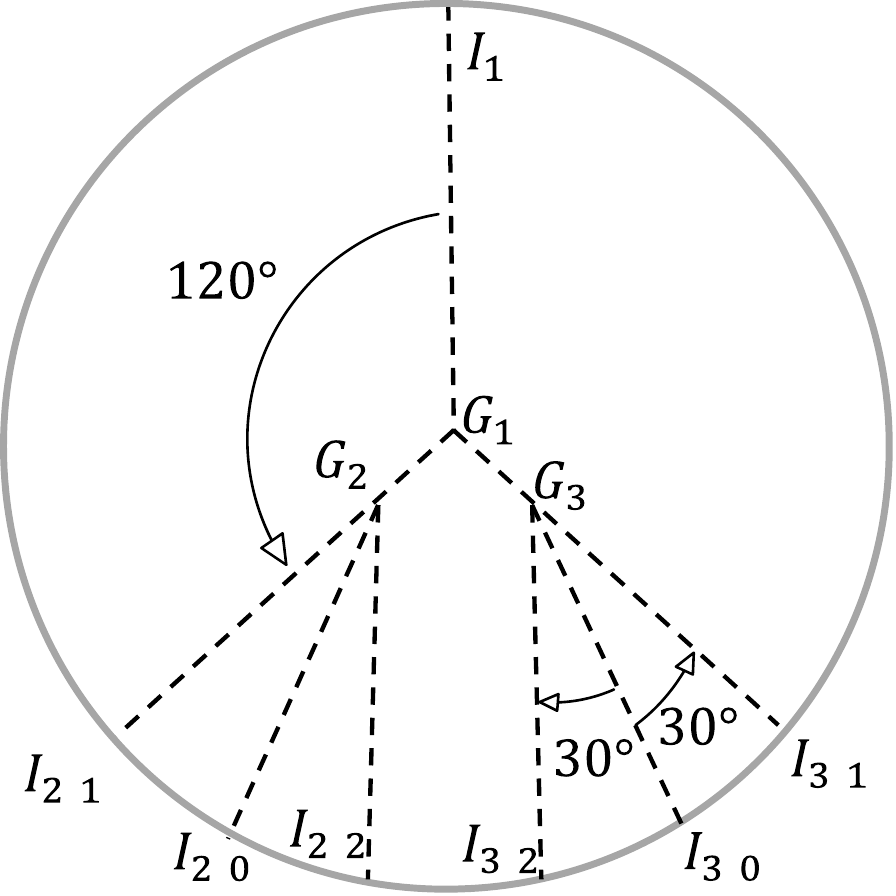}
\caption{Fingers' positions for neutral, cylindrical, spherical grasps}
\label{fig:Grasping}
\end{center}
\end{figure}

Figure \ref{fig:Grasping} shows, in a top view, the ideal positioning of the fingers to switch from prismatic to spherical grips.
The lines $(I_{1} G_{1}),(I_{2-0} G_{2}),(I_{3-0} G_{3})$ represent the orientation of a normal vector on the surface of the distal joint when the three fingers are in the neutral position, the lines  $(I_{1} G_{1}),(I_{2-1} G_{1}),(I_{3-1} G_{1})$  the orientation of a normal vector on the surface of the distal joint when the three fingers are in position to make a cylindrical grasp and the lines $(I_{1} G_{1}),(I_{2-2} G_{2}),(I_{3-2} G_{3})$  the orientation of a normal vector on the surface of the distal joint when the three fingers are in position to make a spherical grasp. Points $G_{1}$, $G_{2}$, $G_{3}$ correspond to the  point $C$ (Figure 
\ref{fig:cinématique_de_la_nouvelle_architecture_de_main}) of each finger. 

\begin{figure}
\begin{center}
	\begin{minipage}[b]{0.15\textwidth}	
	    \center
		\includegraphics[width=3.5cm]{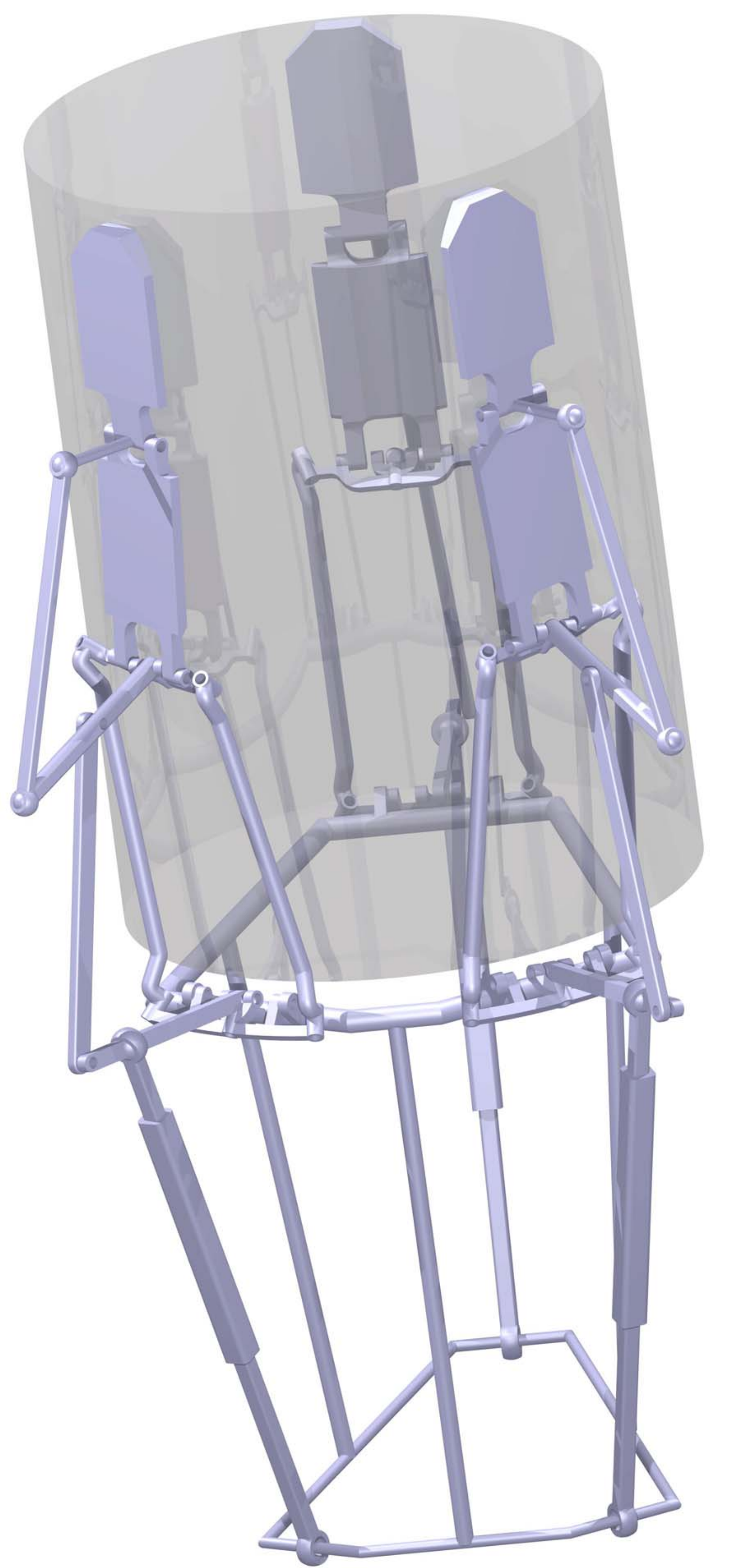}\\
		(a)
	\end{minipage}	
	\hfill
	\begin{minipage}[b]{0.27\textwidth}
	    \center
		\includegraphics[width=4.5cm]{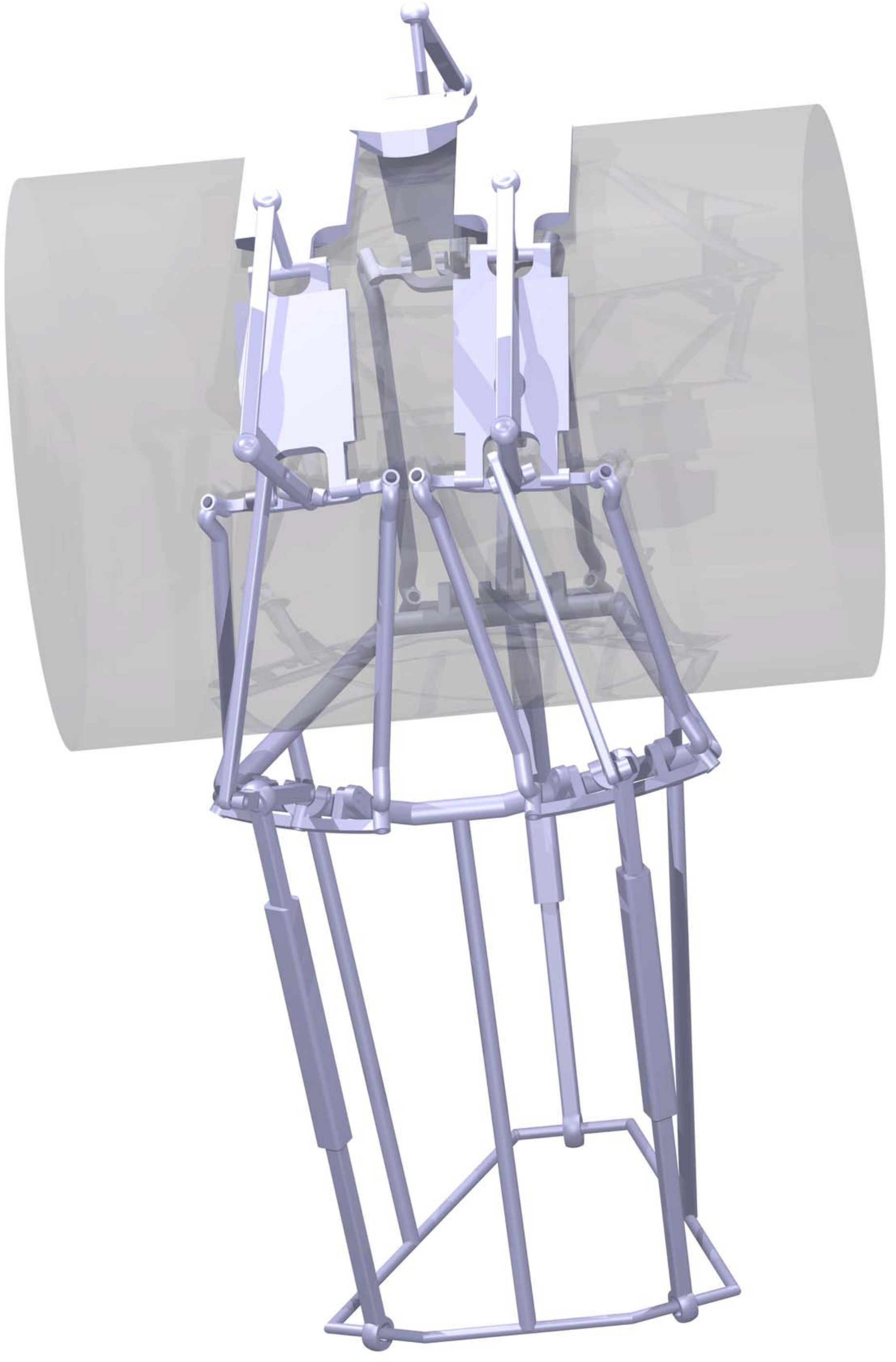}\\
		(b)
	\end{minipage}	
	\caption{(a) Spherical grasp and (b) cylindrical grasp}
	\label{spherical_cylindrical_grip}
\end{center}
\end{figure}

Depending on the shape of the objects to be grasped, the posture of the fingers will be adjusted in a natural way. Springs will allow the fingers to return to a neutral posture in the absence of contact. Figure \ref{spherical_cylindrical_grip}(a) depicts an example of the fingers positions for the spherical grasp and the Fig.~\ref{spherical_cylindrical_grip}(b) the cylindrical grasp.
\section{Stability analysis}
We generalize to this new finger architecture, the under-actuated finger stability theory used for planar fingers \cite{birglen2007underactuated}. The stability criterion is defined as ${\bf f} \geq {\bf 0}$ where $f$ is the vector  of the constraints  of the finger phalanges on the gripping part. The values of ${\bf f}$ is given by 
\begin{equation}
    {\bf f} = {\bf J}^{-t} {\bf T}^{-t} {\bf t}
    \label{f}
\end{equation}
where ${\bf J}$ is the matrix of the torques on the degrees of freedom as a function of the forces applied  on the phalanges, ${\bf T}$ is the matrix of the torque ratios of the transmission mechanism and ${\bf t}$ is the vector of the input force applied by the actuator.

Now the matrices {\bf T} and  {\bf J} are calculated in order to obtain the expression {\bf f}. In the following sub-sections the methodology developed for planar under-actuated fingers \cite{birglen2004kinetostatic} is followed to extend it to the new proposed spatial architecture.
\subsection{Matrix  of the torques {\bf J}}
\begin{figure}
\includegraphics[width=7.5cm]{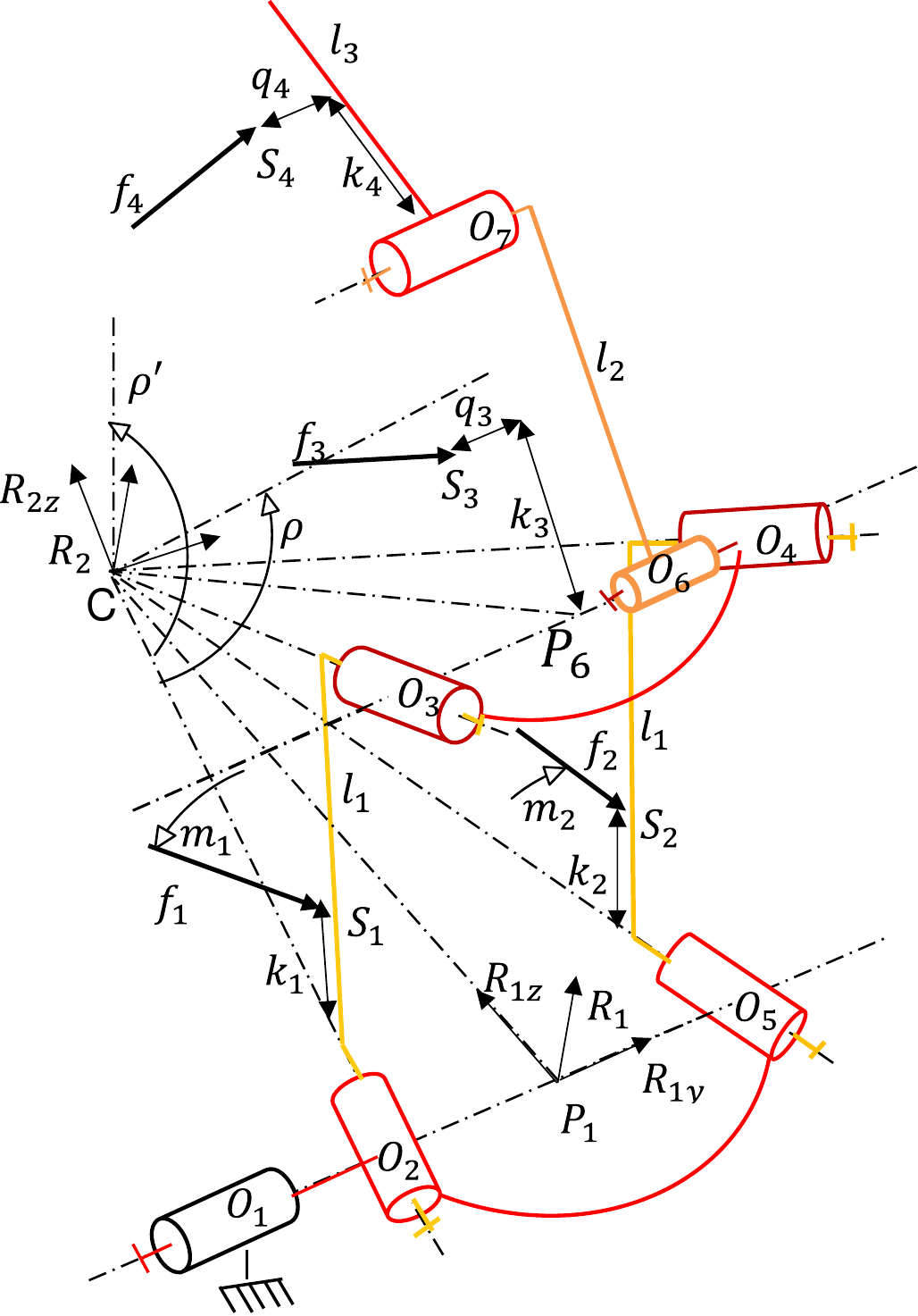}
\caption{Matrix  of the torques {\bf J}}
\label{fig:matrice_J}
\end{figure}

When picking up an object, contacts with the fingers can be located in four places, two at the spherical mechanism, one at the middle phalanx and one at the distal  phalanx.
Define the 4 forces $f_{1}$, $f_{2}$, $f_{3}$, $f_{4}$ (see Figure \ref{fig:matrice_J}) which apply to the points $S_{1}$, $S_{2}$, $S_{3}$, $S_{4}$. $k_{1}$, $k_{2}$, $k_{3}$, $k_{4}$ and $q_{3}$, $q_{4}$ are the distances between these points and the origin of the joints $O_{1}$, $O_{2}$, $O_{6}$, $O_{7}$. $m_{1}$ and $m_{2}$ are the angles between the direction of the forces $f_{1},f_{2}$ and the planes $(O_{2}CO_{3})$ and $(O_{4}CO_{5})$ so that the forces $f_{1}$, $f_{2}$ are parallel to the forces $f_{3}$, $f_{4}$ in the neutral position (home position).

By using spherical trigonometry \cite{todhunter1863spherical}, one gets 
\begin{eqnarray}
m_{1} = \frac{1}{2}\operatorname{acos}{\left(\frac{\cos(\eta)-\sin^{2}{\left(\frac{\alpha}{2} \right)}}
{\cos^{2}{\left(\frac{\alpha}{2} \right)}} \right)}\nonumber\\
m_{2}= -\frac{1}{2}\operatorname{acos}{\left(\frac{\cos(\eta)-\sin^{2}{\left(\frac{\alpha}{2} \right)}}
{\cos^{2}{\left(\frac{\alpha}{2} \right)}} \right)} \label{m}
\end{eqnarray}
To carry out the computations, define a frame located in the middle of point $O_2$ and $O_3$ at point $P_1$ such that  $\overrightarrow{R_{1y}} = \overrightarrow{P_{1}O_{5}} $ and $\overrightarrow{R_{1z}} = \overrightarrow{P_{1}C}$ and a frame located in $C$  such that $\overrightarrow{R_{2z}} = \overrightarrow{O_{2}C} $ and $\overrightarrow{R_{2x}} = \overrightarrow{R_{1x}} $. The matrix {\bf J} depends on the contact points of the finger
\begin{eqnarray}
{\mathbf J} &=& \left[\begin{matrix}\frac{\Theta_{1}}{f_{1}} & \frac{\Theta_{2}}{f_{1}} & 0 & 0\\\frac{\Theta_{1}}{f_{2}} & \frac{\Theta_{2}}{f_{2}} & 0 & 0\\\frac{\Theta_{1}}{f_{3}}& \frac{\Theta_{2}}{f_{3}} &\frac{\Theta_{6}}{f_{3}} & 0\\\frac{\Theta_{1}}{f_{4}} & \frac{\Theta_{2}}{f_{4}} & \frac{\Theta_{6}}{f_{4}} & \frac{\Theta_{7}}{f_{4}}\end{matrix}\right]\label{JJ}
\end{eqnarray}
where $\Theta_{1}$, $\Theta_{2}$, $\Theta_{6}$, $\Theta_{7}$ are the torques along the joint axes $O_{1}$, $O_{2}$, $O_{6}$, $O_{7}$ respectively. The ratios $\frac{\Theta_{7}}{f_{4}},\frac{\Theta_{6}}{f_{4}},\frac{\Theta_{6}}{f_{3}}$ are calculated in the same way as in the planar case \cite{birglen2007underactuated}:
\begin{equation}
    \frac{\Theta_{7}}{f_{4}} = k_{4} \quad
    \frac{\Theta_{6}}{f_{4}} = k_{4} - l_{2} \cos{\left(\delta_{7} + \theta_{7} \right)} \quad
    \frac{\Theta_{6}}{f_{3}} = k_{3}
\end{equation}
Furthermore, $\frac{\Theta_{2}}{f_{2}}$ and $\frac{\Theta_{2}}{f_{1}}$ can be calculated by using the formulae of the spherical parallel mechanisms \cite{mccarthy2010geometric}
\begin{eqnarray}
 \frac{\Theta_{2}}{f_{1}} &=& k_{1} \sin{\left(m_{1} \right)} \cos{\left(\frac{\alpha}{2} \right)} \nonumber\\
 \frac{\Theta_{2}}{f_{2}} &=& \frac{\Theta_{2}}{\Theta_{5}} k_{2} \sin{\left(m_{2} \right)} \cos{\left(\frac{\alpha}{2} \right)}\\
 \frac{\Theta_{2}}{\Theta_{5}} &=& \frac{\sin{\left(\rho \right)}}{\sin{\left(\rho -\eta \right)}}\nonumber
\end{eqnarray}
with $\rho$ the angle between the line $(O_{2}C)$ and the intersection of planes $(O_{2}CO_{5})$ and $(O_{3}CO_{4})$ \cite{mccarthy2010geometric}.

To evaluate $\frac{\Theta_{2}}{f_{3}}$ and $\frac{\Theta_{2}}{f_{4}}$, we calculate the ratio of the speed at the points of application of the forces to multiply them by their components in the $R_{2}$ reference frame.
\begin{eqnarray}
\frac{\Theta_{2}}{\Theta_{3}} &=& \frac{\dot{\theta_{3}}}{\dot{\theta_{2}}}=  \frac{\sin{\left(\rho' \right)}}{\sin{\left(\rho' - \alpha \right)}}
\end{eqnarray}
where $\rho'$ is the angle between the line $(O_{2}C)$ and the intersection of the planes $(O_{2}CO_{3})$ and $(O_{4}CO_{5})$ \cite{mccarthy2010geometric}. Positions $S_{3},S_{4}$ are calculated in the $R_{2}$ frame with
\begin{eqnarray}
\left[S_{3x}~S_{3y}~S_{3z}~1\right]^t &=& Rot_z(\theta_{2}) Rot_x(\alpha) \nonumber \\
& & Rot_z(\theta_{3}) Rot_x(\frac{\eta}{2}) Trans_z(-z_{1}) Rot_y(\theta_{6})\nonumber \\
& & Trans_z(k_{3}) Trans_y(q_{3}) \left[0~0~0~1\right]^t\\
\left[S_{4x}~S_{4y}~S_{4z}~1\right]^t &=& Rot_z(\theta_{2}) Rot_x(\alpha) \nonumber \\
& &Rot_z(\theta_{3}) Rot_x(\frac{\eta}{2}) Trans_z(-z_{1}) \nonumber \\
& &Rot_y(\theta_{6}) Trans_z(l_{2}) Rot_y(\theta_{7})\nonumber \\
& &Trans_z(k_{4}) Trans_y(q_{4}) \left[0~0~0~1\right]^t
\end{eqnarray}
with, if   $0 <\theta_{2}< \pi$ 
\begin{eqnarray}
\theta_{3} &=&- \operatorname{atan_{2}}{\left(B,A \right)} + \operatorname{acos}{\left(\frac{C}{\sqrt{A^{2} + B^{2}}} \right)} \nonumber\\
{\rm else}&:& \\
\theta_{3} &=& - \operatorname{atan_{2}}{\left(B,A \right)} - \operatorname{acos}{\left(\frac{C}{\sqrt{A^{2} + B^{2}}} \right)}\nonumber 
\end{eqnarray}
\begin{eqnarray}
{\rm with} \quad A &=& \sin^{2}{\left(\alpha \right)} \cos{\left(\eta \right)} \cos{\left(\theta_{2} \right)} - \sin{\left(\alpha \right)} \sin{\left(\eta \right)} \cos{\left(\alpha \right)} \nonumber\\
B &=& \sin^{2}{\left(\alpha \right)} \sin{\left(\theta_{2} \right)} \nonumber\\
C &=&\left(\sin{\left(\alpha \right)} \cos{\left(\eta \right)} - \sin{\left(\eta \right)} \cos{\left(\alpha \right)} \cos{\left(\theta_{2} \right)}\right) \sin{\left(\alpha \right)}\nonumber
\end{eqnarray}
The components of $f_3$ and $f_4$ expressed in $R_{2}$ read as
\begin{eqnarray}
\left[f_{3x}~f_{3y}~f_{3z}~1\right]^t &=& Rot_z(\theta_{2}) Rot_x(\alpha) Rot_z(\theta_{3})\nonumber \\ 
& &Rot_x(\frac{\eta}{2}) Rot_y(\theta_{6}) \left[\begin{matrix}1 & 0 & 0& 0\end{matrix}\right]^t\\
\left[f_{4x}~f_{4y}~f_{4z}~1\right]^t &=& Rot_z(\theta_{2}) Rot_x(\alpha) Rot_z(\theta_{3})\nonumber \\ 
& &Rot_x(\frac{\eta}{2}) Rot_y(\theta_{6}) Rot_y(\theta_{7})\left[\begin{matrix}1 & 0 & 0 & 0\end{matrix}\right]^t
\end{eqnarray}
Finally, the ratios $ \frac{\Theta_{2}}{f_{3}}$ and $\frac{\Theta_{2}}{f_{4}}$ are calculated in $R_{1}$
\begin{eqnarray}
\frac{\Theta_{2}}{f_{3}} &=& \frac{\dot{S}_{3x}}{\dot{\theta_{2}}} f_{3x}+\frac{\dot{S}_{3y}}{\dot{\theta_{2}}} f_{3y}+\frac{\dot{S}_{3z}}{\dot{\theta_{2}}} f_{3z}\\
\frac{\Theta_{2}}{f_{4}} &=& \frac{\dot{S}_{4x}}{\dot{\theta_{2}}} f_{4x}+\frac{\dot{S}_{4y}}{\dot{\theta_{2}}} f_{4y}+\frac{\dot{S}_{4z}}{\dot{\theta_{2}}} f_{4z}
\end{eqnarray}
whereas the ratios $ \frac{\Theta_{1}}{f_{3}}$ and $\frac{\Theta_{1}}{f_{4}}$ read as (in $R_{1}$)
\begin{equation}
\frac{\Theta_{1}}{f_{3}} = (S_{3x} f_{3z}-S_{3z} f_{3x})\quad
\frac{\Theta_{1}}{f_{4}} = (S_{4x} f_{4z}-S_{4z} f_{4x})
\end{equation}
The ratios $\frac{\Theta_{1}}{f_{1}}$ and $\frac{\Theta_{1}}{f_{2}}$ are given by (with the positions $S_{1}$, $S_{2}$  given in the reference frame $R_{1}$)
\begin{equation}
\frac{\Theta_{1}}{f_{1}} = (S_{1x} f_{1z}-S_{1z} f_{1x})\quad
\frac{\Theta_{1}}{f_{2}} = (S_{2x} f_{2z}-S_{2z} f_{2x})
\end{equation}
with
\begin{eqnarray}
f_{1x} &=& \sin{\left(m_{1} \right)} \cos{\left(\theta_{2} \right)} + \sin{\left(\theta_{2} \right)} \cos{\left(\frac{\alpha}{2} \right)} \cos{\left(m_{1} \right)}\nonumber\\
f_{1z} &=& \left(\sin{\left(\frac{\alpha}{2} \right)} \cos{\left(\frac{\eta}{2} \right)} + \sin{\left(\frac{\eta}{2} \right)} \cos{\left(\frac{\alpha}{2} \right)} \cos{\left(\theta_{2} \right)}\right) \cos{\left(m_{1} \right)} \nonumber \\
&-& \sin{\left(\frac{\eta}{2} \right)} \sin{\left(m_{1} \right)} \sin{\left(\theta_{2} \right)}\nonumber\\
S_{1x} &=& - k_{1} \sin{\left(\frac{\alpha}{2} \right)} \sin{\left(\theta_{2} \right)}\nonumber\\
S_{1z} &=& k_{1} \left(- \sin{\left(\frac{\alpha}{2} \right)} \sin{\left(\frac{\eta}{2} \right)} \cos{\left(\theta_{2} \right)} + \cos{\left(\frac{\alpha}{2} \right)} \cos{\left(\frac{\eta}{2} \right)}\right) \nonumber\\
f_{2x} &=& \sin{\left(m_{2} \right)} \cos{\left(\theta_{5} \right)} + \sin{\left(\theta_{5} \right)} \cos{\left(\frac{\alpha}{2} \right)} \cos{\left(m_{2} \right)}\nonumber\\
f_{2z} &=& \left(\sin{\left(\frac{\alpha}{2} \right)} \cos{\left(\frac{\eta}{2} \right)} - \sin{\left(\frac{\eta}{2} \right)} \cos{\left(\frac{\alpha}{2} \right)} \cos{\left(\theta_{5} \right)}\right) \cos{\left(m_{2} \right)} \nonumber \nonumber\\
&+& \sin{\left(\frac{\eta}{2} \right)} \sin{\left(m_{2} \right)} \sin{\left(\theta_{5} \right)}\nonumber\\
S_{2x} &=& - k_{2} \sin{\left(\frac{\alpha}{2} \right)} \sin{\left(\theta_{5} \right)}\nonumber\\
S_{2z} &=& k_{2} \left(\sin{\left(\frac{\alpha}{2} \right)} \sin{\left(\frac{\eta}{2} \right)} \cos{\left(\theta_{5} \right)} + \cos{\left(\frac{\alpha}{2} \right)} \cos{\left(\frac{\eta}{2} \right)}\right)\nonumber
\end{eqnarray}
\begin{eqnarray}
{\rm with,~if~} 0 <\theta_{2}:~
\theta_{5} &=&- \operatorname{atan_{2}}{\left(B,A \right)} - \operatorname{acos}{\left(\frac{C}{\sqrt{A^{2} + B^{2}}} \right)} \nonumber\\
{\rm else}&:&\nonumber \\
\theta_{5} &=& - \operatorname{atan_{2}}{\left(B,A \right)} + \operatorname{acos}{\left(\frac{C}{\sqrt{A^{2} + B^{2}}} \right)} \nonumber
\end{eqnarray}
\subsection{Matrix of torques ratios ${\mathbf T}$}
The behaviour of the actuating mechanism is given by the matrix ${\bf T}$, which can be written as follows
The actuation mechanism consists of four loops:
\begin{itemize}
    \item [$\bullet$] A loop ($Loop_{1}$) with five revolute joints ($O_2$, $O_3$, $P_6$, $V_1$, $P_1$). As input, this loop has two degrees of freedom But only one if we consider the spherical mechanism. This {DOF} can be modelled according to $\theta_{2}$. The output of the mechanism can be modelled according to  $\psi_1$, $\nu_1$, $\psi_6$. $V_1$ is a virtual link introduced in section~\ref{V1}.
    We therefore calculate three torque ratios $\frac{\Gamma_{1}}{\Theta_{2}}$, $\frac{N_{1}}{\Theta_{2}}$ and     $\frac{\Gamma_{6}}{\Theta_{2}}$.
    \item [$\bullet$] A loop ($Loop_{2}$) with four revolute joints ($P_5$, $P_6$, $O_6$, $O_7$) and two ball and socket joints  ($P_8$, $P_7$). As input, this loop has three degrees of freedom mechanism that can be modelled according to $\theta_{7},\psi_{6},\theta_{6}$. The output of the mechanism can be modelled according to $\psi_5$.  We therefore calculate three torque ratios $\frac{\Gamma_{5}}{\Theta_{7}}$, $\frac{\Gamma_{5}}{\Gamma_{6}}$ and $\frac{\Gamma_{5}}{\Theta_{6}}$.
    \item [$\bullet$] A five-bar planar linkage ($Loop_{3}$) ($P_2$, $P_3$, $P_4$, $P_5$, $V_1$) which is connected to the spherical mechanism by two revolute joints $P_1$ and $P_6$ which are concurrent in the center of the spherical mechanism $C$ and inscribed in the plane perpendicular to the links of the 5-bar planar mechanism. The joints ($P_1$,$P_2$) and ($P_5$,$P_6$) are equivalent to an universal joint.
    This loop has two degrees of freedom that can be modelled according to $\nu_{1},\psi_{5}$. The output of the mechanism can be modelled according to $\psi_2$.
    We calculate two torque ratios $\frac{\Gamma_{2}}{N_{1}}$ and $\frac{\Gamma_{2}}{\Gamma_{5}}$
    \item [$\bullet$] An actuated prismatic joint $P_{10}$ connect by two ball and socket joint  ($P_{9}$, $P_{11}$) and three revolutes ($O_1$, $P_1$, $P_2$) ($Loop_{4}$). As input, this loop has three degrees of freedom mechanism that can be modelled according to $\theta_{2},\psi_{1},\psi_{2}$. The output of the mechanism is prismatic joint $P_10$.
    We therefore calculate three force ratios $\frac{F_{10}}{\Theta_{1}}$, $\frac{F_{10}}{\Gamma_{1}}$ and $\frac{F_{10}}{\Gamma_{2}}$.
\end{itemize}
\subsubsection{Virtual 2-{DOF}s mechanism ($Loop_{1}$)}
A virtual mechanism with two degrees of freedom ($\nu_1$, $\psi_1$) is defined $(P_{1}, V_{1}, P_{6})$ \label{V1} to model the behaviour of the spherical mechanism on the first stage of the finger actuating mechanism $(P_{2}, P_{3}, P_{4}, P_{5})$. The link $V_{1}$ is defined as coincident at point $C$ and perpendicular to the plane formed by the points $P_{1}, C, P_{6}$.
The virtual mechanism is connected to the point $P_6$ located on the spherical mechanism. 
\begin{figure}
    \center
    \includegraphics[width=7cm]{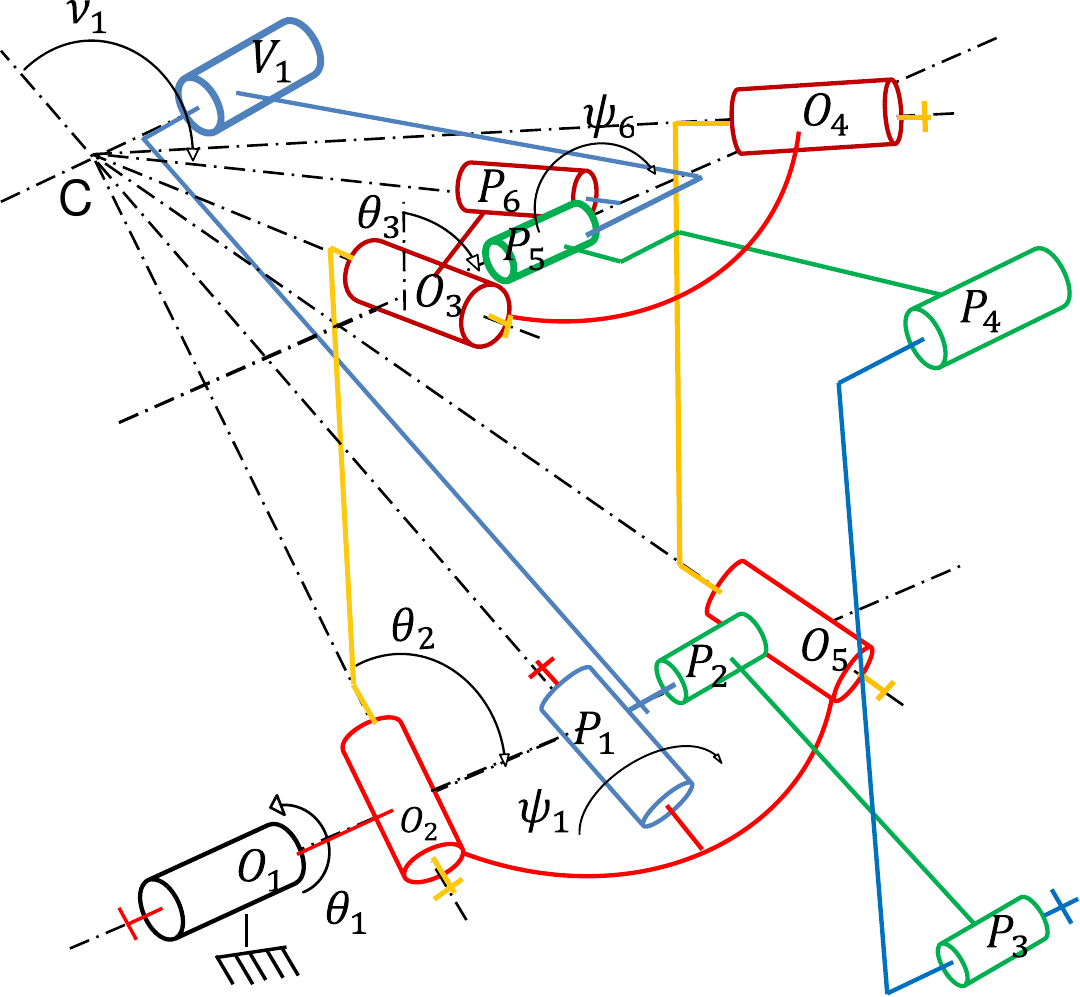}
    \caption{Virtual two-degree-of-freedom mechanism similar to the parallel-spherical mechanism}
    \label{fig:mécanisme_virtuel}
\end{figure}
Depending on the values of $\theta_{2}$, the spherical parallel mechanism admits two assembly modes, parallelogram mode and anti-parallelogram mode. The first mode is used in the finger movement and limits on the passive joints will be placed to avoid the singularity of the mechanism and the change of assembly mode. The direct kinematics of the virtual mechanism is defined as 
\begin{eqnarray}
\nu_{1} &=& \operatorname{acos}{\left(\frac{- P_{6y} \sin{\left(\frac{\eta}{2} \right)} + P_{6x} \cos{\left(\frac{\eta}{2} \right)}}{\cos{\left(\frac{\eta}{2} \right)}} \right)}
\label{nu1}
\end{eqnarray}
\begin{eqnarray}
{\rm If~} P_{6y}>0{\rm~then~}&:& \psi_{1}= -\operatorname{acos}{\left(- \frac{P_{6z}}{\sin{\left(\nu_{1} \right)} \cos{\left(\frac{\eta}{2} \right)}} \right)} \nonumber\\
{\rm else}&:&\nonumber \\
\psi_{1} &=& \operatorname{acos}{\left(- \frac{P_{6z}}{\sin{\left(\nu_{1} \right)} \cos{\left(\frac{\eta}{2} \right)}} \right)}
\label{psi1}
\end{eqnarray}
with $P_6= [P_{6x}, P_{6y}, P_{6z}]^t$  defined as a function of the design parameters and the angles of the spherical parallel mechanism \cite{mccarthy2010geometric}:
\begin{eqnarray}
P_{6x}&=&- \frac{\left(\cos{\left(\eta \right)} + 1\right) \cos{\left(\alpha \right)}}{2} + \frac{\sin{\left(\alpha  \right)} \sin{\left(\eta \right)} \cos{\left(\theta_{3} \right)}}{2} \nonumber\\
P_{6y}&=& \frac{\left(\left(\cos^{2}{\left(\eta \right)}+ 1\right) \sin{\left(\alpha \right)} + \sin{\left(\eta  \right)} \cos{\left(\alpha \right)} \cos{\left(\theta_{3} \right)}\right) \cos{\left(\theta_{2} \right)}}{2} \nonumber\\
&-& \frac{\sin{\left(\eta \right)} \sin{\left(\theta_{2} \right)} \sin{\left(\theta_{3} \right)}}{2}\nonumber\\
P_{6z}&=&- \frac{\left(\left(\cos{\left(\eta \right)} + 1\right) \sin{\left(\alpha \right)} + \sin{\left(\eta \right)} \cos{\left(\alpha \right)} \cos{\left(\theta_{3} \right)}\right) \sin{\left(\theta_{2} \right)}}{2} \nonumber\\
&-& \frac{\sin{\left(\eta \right)} \sin{\left(\theta_{3} \right)} \cos{\left(\theta_{2} \right)}}{2}\nonumber
\end{eqnarray}
Due to the symmetry of the spherical mechanism, we find 
\begin{equation}
\psi_{1} = - \psi_{6}.
\label{psi6}
\end{equation}
The torque ratio $\frac{N_{1}}{\Theta_{2}}$ and $\frac{\Gamma_{1}}{\Theta_{2}}$ can be now written as
\begin{equation}
\frac{N_{1}}{\Theta_{2}} = - \operatorname{asin}{\left(\frac{\dot{P}_{6x} \cos{\left(\frac{\eta}{2} \right)} - \dot{P}_{6y} \sin{\left(\frac{\eta}{2} \right)}}{\cos{\left(\frac{\eta}{2} \right)}} \right)}
\label{N1}
\end{equation}
if $P_{6y}>0$:
\begin{eqnarray}
\frac{\Gamma_{1}}{\Theta_{2}} &=& \frac{\Gamma_{6}}{\Theta_{2}}  = -\frac{N_{1} P_{6z} \cos{\left(\nu_{1} \right)} - \dot{P}_{6z} \sin{\left(\nu_{1} \right)}}{\sin^{2}{\left(\nu_{1} \right)} \cos{\left(\frac{\eta}{2} \right)} \sqrt{- \frac{P_{6z}^{2}}{\sin^{2}{\left(\nu_{1} \right)} \cos^{2}{\left(\frac{\eta}{2} \right)}} + 1} }\\
{\rm else}&:&\nonumber \\
\frac{\Gamma_{1}}{\Theta_{2}} &=& \frac{\Gamma_{6}}{\Theta_{2}}  = \frac{N_{1} P_{6z} \cos{\left(\nu_{1} \right)} - \dot{P}_{6z} \sin{\left(\nu_{1} \right)}}{\sin^{2}{\left(\nu_{1} \right)} \cos{\left(\frac{\eta}{2} \right)} \sqrt{- \frac{P_{6z}^{2}}{\sin^{2}{\left(\nu_{1} \right)} \cos^{2}{\left(\frac{\eta}{2} \right)}} + 1} }
\label{Gamma1}
\end{eqnarray}
based on the velocity of $P_6$:
\begin{eqnarray}
\frac{\dot{P}_{6x}}{\dot{\theta_{2}}} &=& - z_{1} ((- \dot{\theta_{3}} \sin{(\theta_{2} )} \sin{(\theta_{3} )} \cos{(\alpha )} + \dot{\theta_{3}} \cos{(\theta_{2} )} \cos{(\theta_{3} )}  \nonumber \\
&-&\sin{(\theta_{2} )} \sin{(\theta_{3} )} + \cos{(\alpha )} \cos{(\theta_{2} )} \cos{(\theta_{3} )}) \sin{(\frac{\eta}{2} )} \nonumber \\
&+& \sin{(\alpha )} \cos{(\frac{\eta}{2} )} \cos{(\theta_{2} )})\\
\frac{\dot{P}_{6y}}{\dot{\theta_{2}}} &=& - z_{1} ((\dot{\theta_{3}} \sin{(\theta_{2} )} \cos{(\theta_{3} )} + \dot{\theta_{3}} \sin{(\theta_{3} )} \cos{(\alpha )} \cos{(\theta_{2} )}   \nonumber \\
&+&\sin{(\theta_{2} )} \cos{(\alpha )} \cos{(\theta_{3} )} + \sin{(\theta_{3} )} \cos{(\theta_{2} )}) \sin{(\frac{\eta}{2} )} \nonumber \\
&+& \sin{(\alpha )} \sin{(\theta_{2} )} \cos{(\frac{\eta}{2} )})\\
\frac{\dot{P}_{6z}}{\dot{\theta_{2}}} &=& - z_{1} \dot{\theta_{3}} \sin{(\alpha )} \sin{(\frac{\eta}{2} )} \sin{(\theta_{3} )}
\end{eqnarray}
\subsubsection{Evaluation of  $\Gamma_{5}/\Theta_{7}$  ($Loop_{2}$)}
The mechanism formed by the joints ($0_{7}, P_{8}, P_{9}, P_{5}, P_{6}, 0_{6}$) is a three degrees of freedom mechanism that can be modelled according to $\theta_{7},\psi_{6},\theta_{6}$.
The angles $\theta_{7}$, $\theta_{6}$ are degrees of freedom corresponding to the movements of the intermediate and distal phalanges and the angle $\psi{6}$ is related to the motions of the spherical parallel mechanism $\theta_{2}$ by Eqs.~\eqref{psi6} and \eqref{psi1}. $\Theta_{7}$ is the torque in the joint $0_{7}$ and $\Gamma_{5}$ is the torque in the link $P_{5}$, to calculate the torque ratio $\frac{\Gamma_{5}}{\Theta_{7}}$, we define $RSSR$ mechanism with one degree of freedom \cite{mazzotti2017dimensional} with four joints ($0_{7}$, $P_{8}$, $P_{9}$, $P_{5}$). The joints $P_{6}$ and $0_{6}$ are defined in the matrix $M_{O75}$ of Eq.~\eqref{MO75} of transfer between links ($0_{7}$, $P_{5}$).
\begin{figure}
    \center
    \includegraphics[width=7cm]{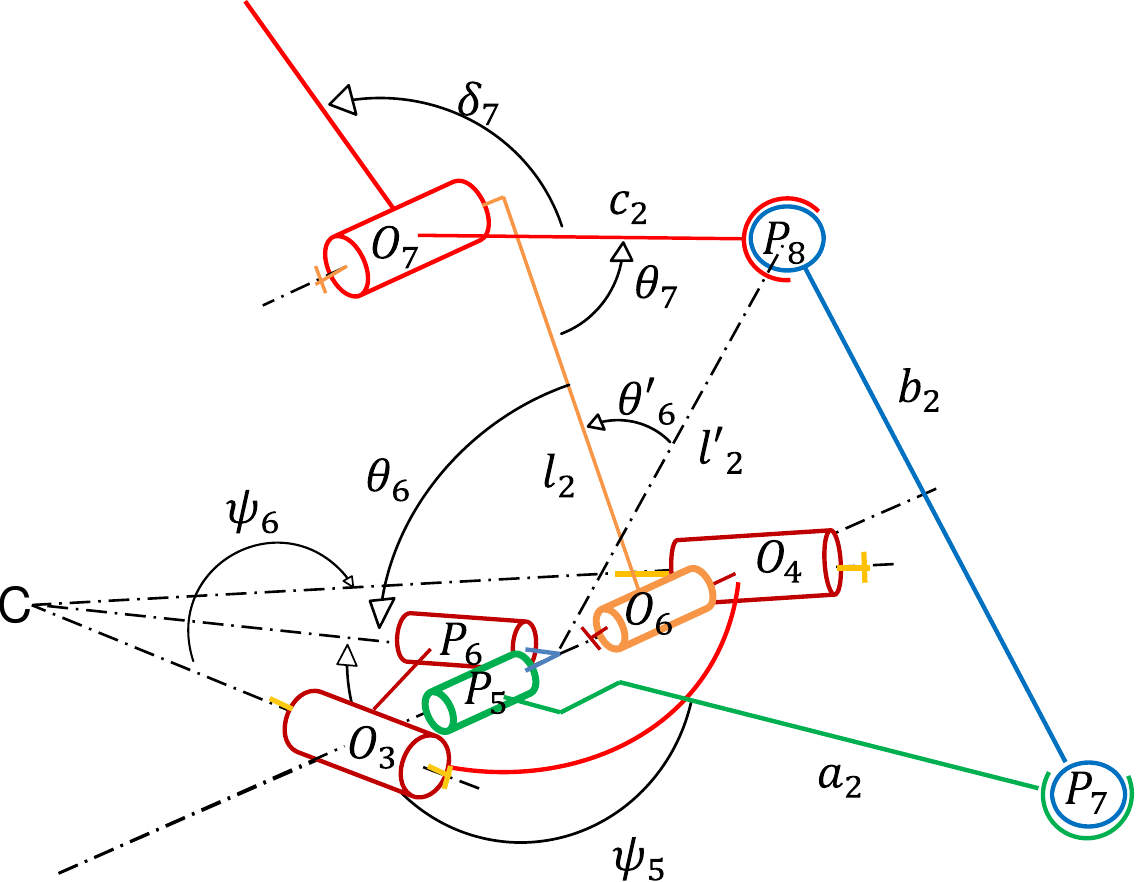}
    \caption{Mechanism $RSSR$}
\label{fig:système_RSSR}
\end{figure}

\begin{eqnarray}
M_{O75}&=& \left[\begin{matrix}s_{11} & s_{12} & s_{13} & X_{01}\\s_{21} & s_{22} & s_{23} & Y_{01}\\s_{31} & s_{32} & s_{33} & Z_{01}\\0 & 0 & 0 & 1\end{matrix}\right]\label{MO75}\\
&=& \left[\begin{matrix}- \cos{\left(\theta_{6} \right)} & - \sin{\left(\theta_{6} \right)} \cos{\left(\psi_{6} \right)} & \sin{\left(\psi_{6} \right)} \sin{\left(\theta_{6} \right)} & l_{2}\\\sin{\left(\theta_{6} \right)} & - \cos{\left(\psi_{6} \right)} \cos{\left(\theta_{6} \right)} & \sin{\left(\psi_{6} \right)} \cos{\left(\theta_{6} \right)} & 0\\0 & \sin{\left(\psi_{6} \right)} & \cos{\left(\psi_{6} \right)} & 0\\0 & 0 & 0 & 1\end{matrix}\right]\nonumber 
\end{eqnarray}
The torque ratio $\frac{\Gamma_{5}}{\Theta_{7}}$ can be now written as
\begin{equation}
\frac{\Gamma_{5}}{\Theta_{7}} = \frac{d' \cos{\left(\psi_{5} \right)} + e' \sin{\left(\psi_{5} \right)} - f'}{d \sin{\left(\psi_{5} \right)} - e \cos{\left(\psi_{5} \right)}}
\label{moneq}
\end{equation}
with~$d' = 2 a_{2} c_{2} s_{11} \sin{\left(\theta_{7} \right)} - 2 a_{2} c_{2} s_{21} \cos{\left(\theta_{7} \right)}$, ~
$e' = 2 a_{2} c_{2} s_{12} \sin{\left(\theta_{7} \right)} - 2 a_{2} c_{2} s_{22} \cos{\left(\theta_{7} \right)}$,~
$f' =- 2 X_{01} c_{2} \sin{\left(\theta_{7} \right)}$,~ 
and where $\psi_{5}$ is the angle between $\overrightarrow{P_{5}C}$ and $\overrightarrow{P_{5}P_{7}}$ as defined as following
\begin{equation}
\psi_{5} = \operatorname{atan_{2}}{\left(e,d \right)} - \operatorname{acos}{\left(\frac{f}{\sqrt{d^{2} + e^{2}}} \right)} 
\label{psi5}
\end{equation}
with $d = - 2 a_{2} c_{2} s_{21} \sin{\left(\theta_{7} \right)} + 2 a_{2} s_{11} \left(X_{01} - c_{2} \cos{\left(\theta_{7} \right)}\right)$,~ 
$e = - 2 a_{2} c_{2} s_{22} \sin{\left(\theta_{7} \right)} + 2 a_{2} s_{12} \left(X_{01} - c_{2} \cos{\left(\theta_{7} \right)}\right)$ and
$f = - X_{01}^{2} + 2 X_{01} c_{2} \cos{\left(\theta_{7} \right)} - a_{2}^{2} + b_{2}^{2} - c_{2}^{2}$.
\subsubsection{ Evaluation of  $\Gamma_{5}/\Theta_{6}$ ($Loop_{2}$)}
In a similar way, to calculate $\Gamma_{5}$ according to $\Theta_{7}$, we calculate torque ratio $\frac{\Gamma_{5}}{\Theta_{6}}$, where  $\Theta_{6}$ is the torque in the joint $0_{6}$. We define the $RSSR$ mechanism with one  degree of freedom \cite{mazzotti2017dimensional} with for joints ($0_{6},P_{8},P_{9},P_{5}$). The joint $P_{6}$ is defined by the matrix $M_{O65}$ Eq.~\eqref{MO65} of transfer between links $0_{6},P_{5}$.
\begin{equation}
M_{O65} =
\left[\begin{matrix}1 & 0 & 0 & 0\\0 & \cos{\left(\psi_{6} \right)} & - \sin{\left(\psi_{6} \right)} & 0\\0 & \sin{\left(\psi_{6} \right)} & \cos{\left(\psi_{6} \right)} & 0\\0 & 0 & 0 & 1\end{matrix}\right]
\label{MO65}
\end{equation}
We introduce two new variables $\theta'_{6}$, $l'_{2}$ associated with the links ($0_{6}$, $P_{8}$).
\begin{equation}
l'_{2} = \sqrt{c_{2}^{2} - 2 c_{2} l_{2} \cos{\left(\theta_{7} \right)} + l_{2}^{2}}
\label{l'2}
\end{equation}
\begin{eqnarray}
{\rm if~} \pi<\theta_{7}< 2 \pi&:& \nonumber \\
\theta'_{6} &=& \operatorname{acos}{\left(\frac{l_{2}^{2} + (l'_{2})^{2}- c_{2}^{2}}{2 l_{2} l'_{2}} \right)}\nonumber \\
{\rm else}&:&\nonumber \\
\theta'_{6} &=& - \operatorname{acos}{\left(\frac{ l_{2}^{2} + (l'_{2})^{2}- c_{2}^{2}}{2 l_{2} l'_{2}} \right)}
\end{eqnarray}
The torque ratio $\frac{\Gamma_{5}}{\Theta_{6}}$ can be now written as
\begin{equation}
\frac{\Gamma_{5}}{\Theta_{6}} = \frac{- d' \cos{\left(\psi_{5} \right)} - e' \sin{\left(\psi_{5} \right)}}{d \sin{\left(\psi_{5} \right)} - e \cos{\left(\psi_{5} \right)}}
\label{Gamma5Theta6}
\end{equation}
with $d' = 2 a_{2} l'_{2} \sin{\left(\theta_{6} + \theta'_{6} \right)}$,~ $e' = - 2 a_{2} l'_{2} \cos{\left(\psi_{6} \right)} \cos{\left(\theta_{6} + \theta'_{6} \right)}$, ~$d = - 2 a_{2} l'_{2} \cos{\left(\theta_{6} + \theta'_{6} \right)}$ and $e = - 2 a_{2} l'_{2} \cos{\left(\psi_{6} \right)} \sin{\left(\theta_{6} + \theta'_{6} \right)}$.
\subsubsection{ Evaluation of  $\Gamma_{5}/\Gamma_{6}$ ($Loop_{2}$)}
In a similar way, to calculate $\Gamma_{5}$ according to $\Theta_{7}$, we calculate $\frac{\Gamma_{5}}{\Gamma_{6}}$, where  $\Gamma_{6}$ is the torque in the joint $P_{6}$ Eq.~\eqref{Gamma1}.
We define the $RSSR$ mechanism with one  degree of freedom \cite{mazzotti2017dimensional} with for joints ($P_{6},P_{8},P_{9},P_{5}$). The the matrix $M_{P65}$ Eq.~\eqref{MP65} is matrix of transfer between links $P_{6},P_{5}$.
\begin{figure}
    \center
    \includegraphics[width=7cm]{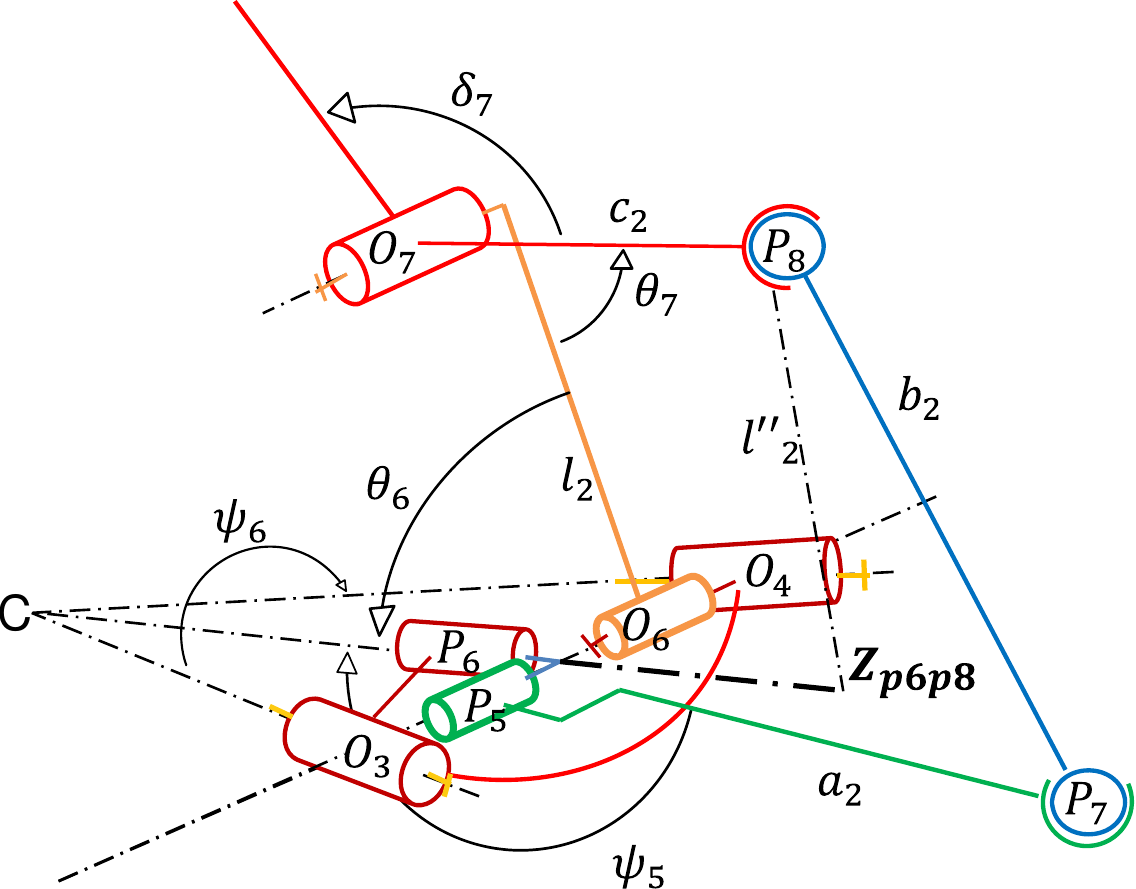}
    \caption{Mechanism $RSSR$ $\Gamma_{5}(\Gamma_{6})$ }
    \label{fig:système_RSSR_G6}
\end{figure}
\begin{eqnarray}
M_{P65} &=&\left[\begin{matrix}0 & -1 & 0 & 0\\0 & 0 & -1 & 0\\1 & 0 & 0 & Z_{P6P8}\\0 & 0 & 0 & 1\end{matrix}\right] \label{MP65}
\end{eqnarray}
We introduce two new variables $Z_{P6P8}$, $l''_{2}$ associated with the links ($P_{6}$, $P_{8}$).
\begin{eqnarray}
l''_{2} = c_{2} \sin{\left(\theta_{6} + \theta_{7} \right)} - l_{2} \sin{\left(\theta_{6} \right)}\\
Z_{P6P8} = c_{2} \sin{\left(\theta_{6} + \theta_{7} \right)} - l_{2} \sin{\left(\theta_{6} \right)}
\label{l''2}
\end{eqnarray}
The torque ratio $\frac{\Gamma_{5}}{\Gamma_{6}}$ can be now written as
\begin{equation}
\frac{\Gamma_{5}}{\Gamma_{6}} = \frac{d e'}{d^{2} + e^{2}} - \frac{d' e' f}{\left(d^{2} + e^{2}\right) \sqrt{d^{2} + e^{2} - f^{2}}}
\label{Gamma65}
\end{equation}
with $d' = 2 a_{2} l''_{2} \cos{\left(\psi_{6} \right)}$, $e' = - 2 a_{2} l''_{2} \sin{\left(\psi_{6} \right)}$, 
~$d = 2 Z_{P6P8} a_{2}$, $e = 2 a_{2} l''_{2} \cos{\left(\psi_{6} \right)}$ and 
~$f = - Z_{P6P8}^{2} - a_{2}^{2} + b_{2}^{2} - (l''_{2})^{2}$.
\subsubsection{Evaluation of  $\Gamma_{2}/\Gamma_{5}$ and $\Gamma_{2}/N_{1}$ ($Loop_{3}$)}
\begin{figure}
    \center
    \includegraphics[width=7cm]{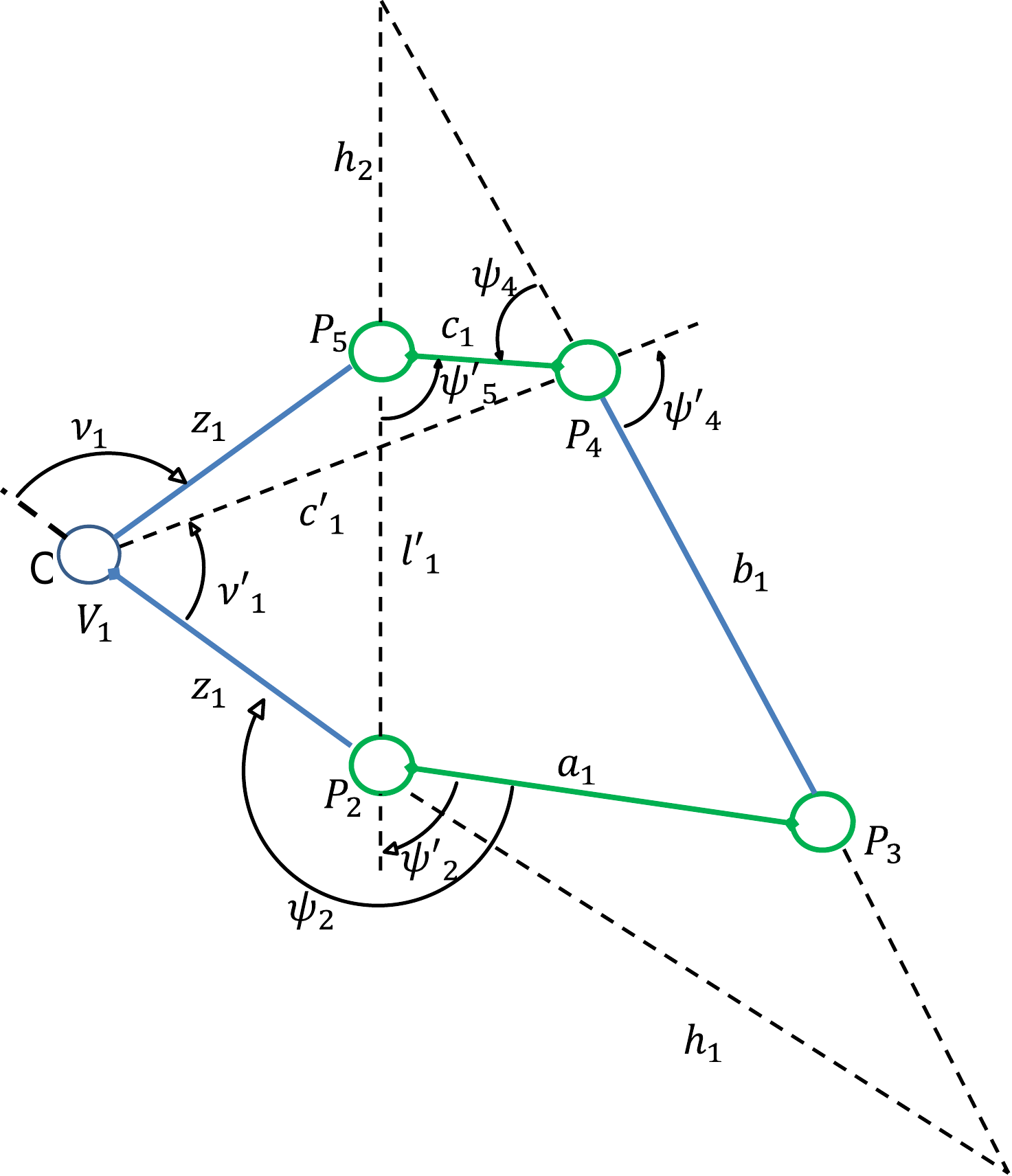}
    \caption{Actuation coupling mechanism}
    \label{fig:système_plan}
\end{figure}

The joints ($P_{5}$, $P_{4}$, $P_{3}$, $P_{2}$, $V_{1}$) form a closed loop with two degrees of freedom (Figure~\ref{fig:système_plan}). $\nu_{1}$ is linked to the movements of the spherical mechanism $\theta_{2}$ by Eq.~\eqref{nu1} and $\psi_{5}$ is calculated in Eq.~\eqref{psi5}. To calculate the torque ratio $\frac{\Gamma_{2}}{\Gamma_{5}}$, we define the 4R mechanism \cite{mccarthy2010geometric} plane with for the articulations ($P_{5},P_{4},P_{3},P_{2}$).

To do this, we define three new variables: $l'_{1}$ distance between $P_{5},P_{2}$,  $\psi'_{5}$ the angle between $\overrightarrow{P_{5}P_{7}}$ and $\overrightarrow{P_{5}P_{2}}$ and $z_{1}$ distance between C$P_{1}$ and C$P_{5}$ 
\begin{equation}
z_{1} = \frac{l_{1} \cos{\left(\frac{\eta}{2} \right)}}{2 \sin{\left(\frac{\alpha}{2} \right)}}
\label{moneq4}
\end{equation}
We deduce from this
\begin{equation}
l'_{1}= 2 z_{1} \cos{\left(\frac{\nu_{1}}{2} \right)}
\label{moneq5}
\end{equation}
with $\nu_{1}$ defined in Eq.~\eqref{nu1}
\begin{equation}
\psi'_{5} = - \frac{\nu_{1}}{2} - \psi_{5}
\label{moneq6}
\end{equation}
The angles $\psi_{4}$ are written as a function of $\psi'_{5}$
\begin{equation}
\psi_{4}(\psi'_{5}) =  \operatorname{atan_{2}}{\left(B,A \right)}- \operatorname{acos}{\left(\frac{C}{\sqrt{A^{2} + B^{2}}} \right)} 
\label{moneq7}
\end{equation}
with $A(\psi'_{5}) = 2 b_{1} c_{1} - 2 b_{1} l'_{1} \cos{\left(\psi'_{5} \right)}$, $B(\psi'_{5}) = 2 b_{1} l'_{1} \sin{\left(\psi'_{5} \right)}$ and
$C(\psi'_{5}) = a_{1}^{2} - b_{1}^{2} - c_{1}^{2} + 2 c_{1} l'_{1} \cos{\left(\psi'_{5} \right)} - (l'_{1})^{2}$.

The torque ratio $\frac{\Gamma_{2}}{\Gamma_{5}}$ can be now written as
\begin{equation}
\frac{\Gamma_{2}}{\Gamma_{5}} = \frac{h_{2}}{h_{2} + l'_{1}}
\end{equation}
\begin{equation}
{\rm with} \quad h_{2} = - \frac{c_{1} \sin{\left(\psi_{4} \right)}}{\sin{\left(\psi_{4} + \psi'_{5} \right)}} \nonumber
\end{equation}

To calculate the torque ratio $\frac{\Gamma_{2}}{N_{1}}$, we define the 4R mechanism \cite{mccarthy2010geometric} plane with for the articulations ($V_{1},P_{4},P_{3},P_{2}$). To do this, we define three new variables: $c'_{1}$ distance between $V_{1},P_{4}$,  $\nu'_{1}$ the angle between $\overrightarrow{V_{1}P_{4}}$ and $\overrightarrow{V_{1}P_{2}}$ and $\psi'_{4}$ the angle between $\overrightarrow{P_{4}V_{1}}$ and $\overrightarrow{P_{4}P_{3}}$
\begin{eqnarray}
c'_{1}&=& \sqrt{c_{1}^{2} - 2 c_{1} z_{1} \cos{\left(\psi_{6} \right)} + z_{1}^{2}} \label{c'1} \\
%
\nu'_{1}&=& - \nu_{1} - \operatorname{acos}{\left(\frac{- c_{1}^{2} + (c'_{1})^{2} + z_{1}^{2}}{2 c'_{1} z_{1}} \right)} + \pi
\label{nu'1}
\end{eqnarray}
The angle $\psi'_{4}$ as a function of $\nu'_{1}$ is defined
\begin{equation}
\psi'_{4}(\nu'_{1}) = \operatorname{atan_{2}}{\left(B,A \right)}- \operatorname{acos}{\left(\frac{C}{\sqrt{A^{2} + B^{2}}} \right)} 
\label{psi'4}
\end{equation}
with $A(\nu'_{1}) = 2 b_{1} c'_{1} - 2 b_{1} z_{1} \cos{\left(\nu'_{1} \right)}$, 
$B(\nu'_{1}) = 2 b_{1} z_{1} \sin{\left(\nu'_{1} \right)}$ and
$C(\nu'_{1}) = a_{1}^{2} - b_{1}^{2} - (c'_{1})^{2} + 2 c'_{1} z_{1} \cos{\left(\nu'_{1} \right)} - z_{1}^{2}$.

\begin{equation}
h_{1} = - \frac{c'_{1} \sin{\left(\psi'_{4} \right)}}{\sin{\left(\nu'_{1} + \psi'_{4}\right)}}
\label{8bis}
\end{equation}
The torque ratio $\frac{\Gamma_{2}}{N_{1}}$ can be now written as
\begin{equation}
\frac{\Gamma_{2}}{N_{1}} = - \frac{h_{1}}{h_{1} + z_{1}}
\label{moneq8ter}
\end{equation}
The angles $\psi_{2}$ and $\psi'_{2}$ are calculated to allow the calculation of the force on the actuator. 
\begin{eqnarray}
\psi_{2}(\psi'{5}) &=&  \psi'_{2}(\psi'_{5}) - \frac{\nu_{1}}{2} \\
\psi'_{2}(\psi'_{5}) &=& \operatorname{atan}{\left(\frac{B}{A} \right)} - \operatorname{acos}{\left(\frac{C}{\sqrt{A^{2} + B^{2}}} \right)}
\label{moneq9}
\end{eqnarray}
with $A(\psi'_{5}) = 2 a_{1} c_{1} \cos{\left(\psi'_{5} \right)} - 2 a_{1} l_{1}$, 
$B(\psi'_{5}) = 2 a_{1} c_{1} \sin{\left(\psi'_{5} \right)}$ and
$C(\psi'_{5}) = a_{1}^{2} - b_{1}^{2} + c_{1}^{2} - 2 c_{1} l_{1} \cos{\left(\psi'_{5} \right)} + l_{1}^{2}$.
\subsubsection{Actuating loop ($Loop_{4}$)}
Finger actuation is done with the prismatic link $P_{10}$ which is connected to two ball-and-socket joints $P_{9}$ and $P_{11}$. The distance between $P_{9}$ and $P_{11} $ is named $d_{0}$ 
, where $a_{0}$ and $c_{0} $ are the lengths between $P_{9}$ and $O_{1} $, $P_{11}$ and $0_{1} $, respectively.
\begin{multline}
d_{0} = \\
\sqrt{a_{0}^{2} + 2 a_{0} c_{0} (\sin{\left(\psi_{2} \right)} \sin{\left(\theta_{1} \right)} \cos{\left(\psi_{1} \right)} + \cos{\left(\psi_{2} \right)} \cos{\left(\theta_{1} \right)}) + c_{0}^{2}} \nonumber
\end{multline}
We derive $d_{0}$ according to $\psi_{2}$, $\psi_{1}$ and $\theta_{1}$ to obtain $\frac{F_{10}}{\Gamma_{2}}, \frac{F_{10}}{\Gamma_{1}}, \frac{F_{10}}{\Theta_{1}}$ where $F_{10}$ is the force produced by the actuator.
\begin{eqnarray}
\frac{F_{10}}{\Gamma_{2}} &=&  \frac{a_{0} c_{0} \left(\sin{\left(\psi_{2} \right)} \cos{\left(\theta_{1} \right)} - \sin{\left(\theta_{1} \right)} \cos{\left(\psi_{1} \right)} \cos{\left(\psi_{2} \right)}\right) }{d_{0}} \label{moneq_gamma2} \nonumber\\
\frac{F_{10}}{\Gamma_{1}} &=&  - \frac{a_{0} c_{0} \sin{\left(\psi_{1} \right)} \sin{\left(\psi_{2} \right)} \sin{\left(\theta_{1} \right)}}{d_{0}}
\label{moneq_gamma1}\nonumber \\
\frac{F_{10}}{\Theta_{1}} &=&  \frac{a_{0} c_{0} \left(\sin{\left(\psi_{2} \right)} \cos{\left(\psi_{1} \right)} \cos{\left(\theta_{1} \right)} - \sin{\left(\theta_{1} \right)} \cos{\left(\psi_{2} \right)}\right)}{d_{0}}\nonumber
\label{moneq_theta1}
\end{eqnarray}
\begin{figure}
    \center
    \includegraphics[width=6.5cm]{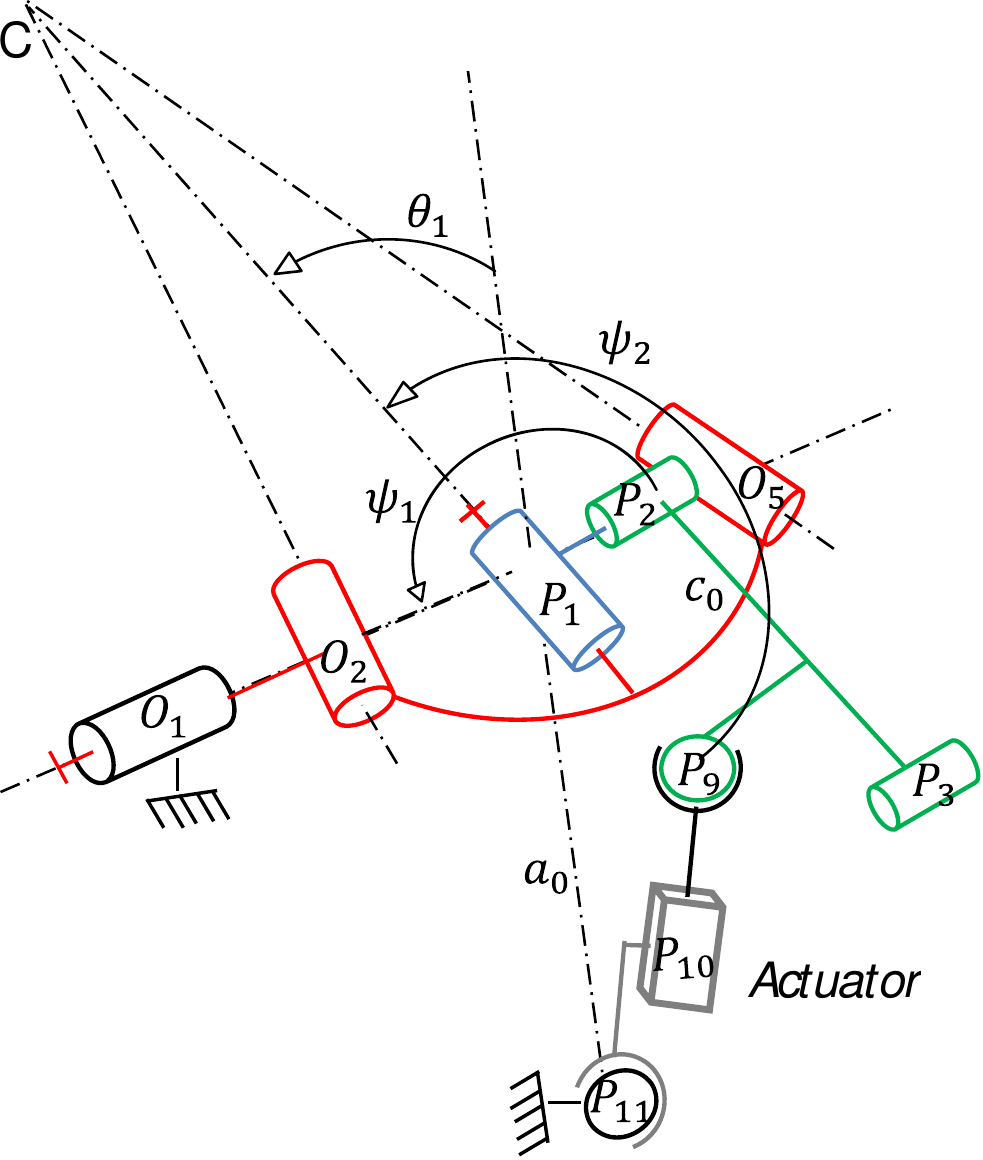}
    \caption{Actuating loop ($Loop_{4}$)}
    \label{fig:actuator_system}
\end{figure}
\subsubsection{Evaluation of matrix {\bf T}}  
From the previous results, it is now possible to write the matrix ${\bf T}$ as follows
\begin{equation}
{\bf T} = \left[\begin{matrix}-\frac{F_{10}}{\Theta_{1}} & -\frac{F_{10}}{\Theta_{2}} & -\frac{\Gamma_{2}}{\Theta_{6}} \frac{\Gamma_{2}}{\Gamma_{5}} \frac{F_{10}}{\Gamma_{2}} & -\frac{\Gamma_{5}}{\Theta_{7}} \frac{\Gamma_{2}}{\Gamma_{5}} \frac{F_{10}}{\Gamma_{2}}\\0 & 1 & 0 & 0\\0 & 0 & 1 & 0\\0 & 0 & 0 & 1\end{matrix}\right]
\end{equation}
where $\quad \frac{F_{10}}{\Theta_{2}} = \left(\frac{\Gamma_{1}}{\Theta_{2}} \frac{F_{10}}{\Gamma_{1}} + \frac{N_{1}}{\Theta_{2}} \frac{\Gamma_{2}}{N_{1}} \frac{F_{10}}{\Gamma_{2}} + \frac{\Gamma_{6}}{\Theta_{2}} \frac{\Gamma_{5}}{\Gamma_{6}} \frac{\Gamma_{2}}{\Gamma_{5}} \frac{F_{10}}{\Gamma_{2}}\right)$  
consists of three terms because when the parallel spherical mechanism moves, it influences all the loops of the mechanism.
\subsection{Input-output analysis for stability analysis}
\begin{figure}
\includegraphics[width=10cm]{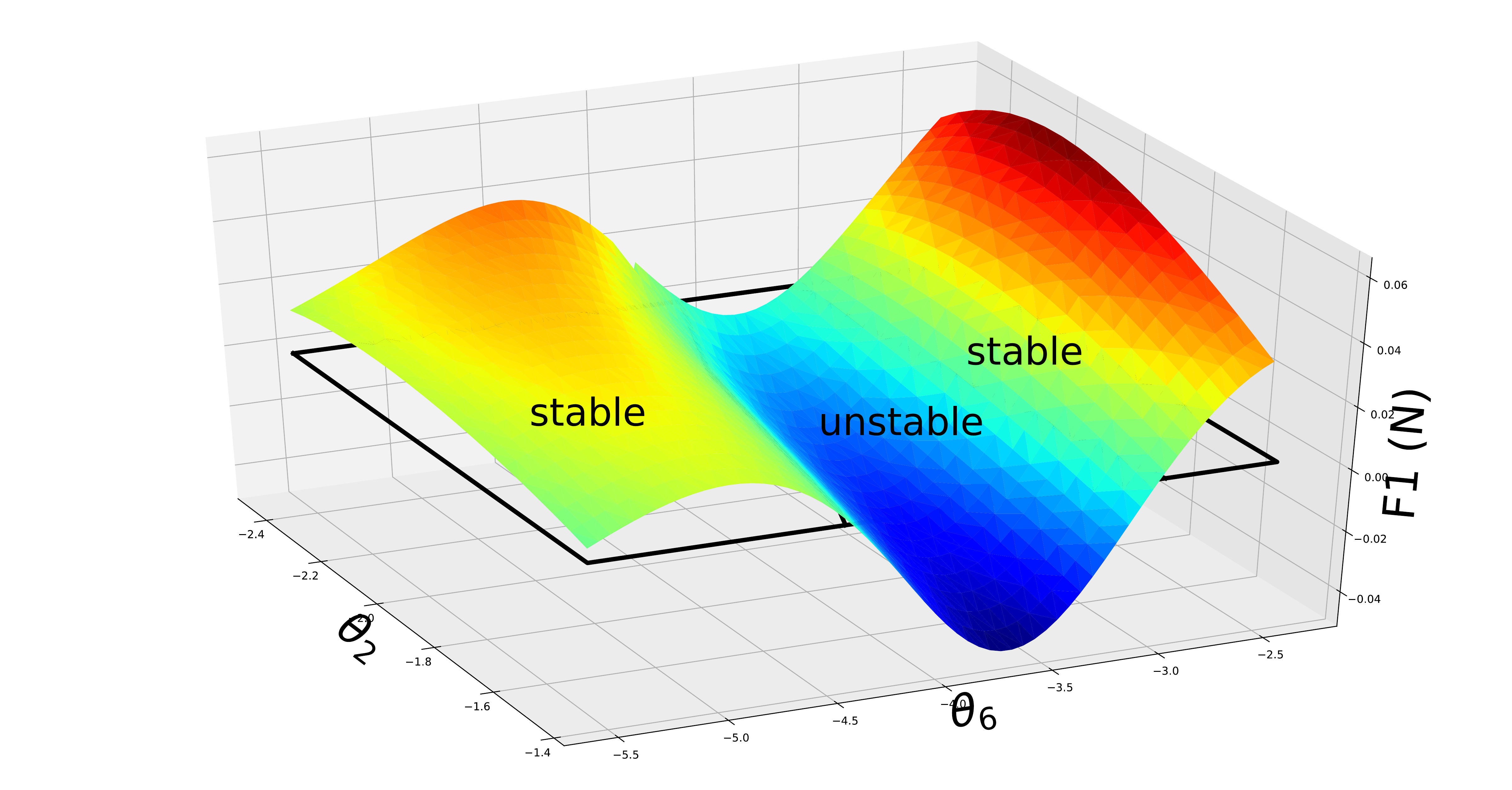}
\caption{Stability condition for $f_1$ (N) versus $\theta_2$ and $\theta_6$ [rad].}
\label{fig:F1}
\end{figure}

The input force ${\bf t} = \left[f_{10}~0~0~0\right]^t\label{J}$ is defined by the action forces with the return springs are neglected and the output forces ${\bf f} = \left[f_{1}~f_{2}~f_{3}~f_{4}\right]^t$
are defined by the contact forces applied by the finger to the object . 
Stable configurations can be found in Fig.~\ref{fig:F1} with Tab.~\ref{tab1}. The movement $\theta_{2}$ causes limited variation in strength $f_{1}$ where $\theta_{1}$, $\theta_{7}$ (See Eq.~\eqref{theta7}) and $m_{1}$, $m_{2}$ (See Eq.~\eqref{m}) are in home position.
\begin{table}
    \caption{Design parameters}
    \label{tab1}
    \begin{tabular}{|*{8}{c|}}
        \hline
        $l_{1}$ & $l_{2}$ & $l_{3}$ & $a_{0}$ & $a_{123}$ & $b_{i}$ & $c_{0}$ & $c_{123}$ \\
        \hline
        61 & 41 & 38 & 100 & 38 & 58 & 28 & 16\\
        \hline
        $k_{12}$ & $k_{3}$ & $k_{4}$ & $q_{i}$ & $\delta_{7}$ & $\alpha$ & $\eta$ & $f_{10}$ \\
        \hline
        $l_{1}/2$ & $l_{2}/2$ & $l_{3}/2$ & 0 & $\pi/2$ & 85deg & 40deg & 1 N\\
        \hline
    \end{tabular}
\end{table}
\section{Conclusions}
The main contribution of this article is a new kinematics allowing complex grasps with a spatial under-actuated mechanism. A spherical parallel mechanism has been added to the classical under-actuated fingers in order to adjust the fingers from a neutral position to a spherical or cylindrical grasp. The stability of this new robotic hand was analysed using two Jacobian matrices.  Compared to the modelling of planar under-actuated fingers, new components have been added to the Jacobian matrix. The contact surfaces are simplified by point contacts.
In this article, the interaction between the different fingers to allow a stable grip is not studied and will be the subject of further work. Future work will also be carried out to optimize the design parameters of the fingers for a family of parts from industry.
\section*{Acknowledgements}
This work comes from a PhD thesis granted by ANRT (CIFRE program).
\bibliographystyle{asmems4}
\bibliography{asme2e}

\begin{thebibliography}{10}

\bibitem{okada1982computer}
Okada, T., 1982.
\newblock ``Computer control of multijointed finger system for precise
  object-handling''.
\newblock {\em IEEE Transactions on Systems, Man, and Cybernetics, \textbf{
  12}}(3), pp.~289--299.

\bibitem{salisbury1982articulated}
Salisbury, J.~K., and Craig, J.~J., 1982.
\newblock ``Articulated hands: Force control and kinematic issues''.
\newblock {\em The International journal of Robotics research, \textbf{ 1}}(1),
  pp.~4--17.

\bibitem{jacobsen1986design}
Jacobsen, S., Iversen, E., Knutti, D., Johnson, R., and Biggers, K., 1986.
\newblock ``Design of the utah/mit dextrous hand''.
\newblock In Proceedings. 1986 IEEE International Conference on Robotics and
  Automation, Vol.~3, IEEE, pp.~1520--1532.

\bibitem{gazeau2001lms}
Gazeau, J.-P., Zehloul, S., Arsicault, M., and Lallemand, J.-P., 2001.
\newblock ``The lms hand: force and position controls in the aim of the fine
  manipulation of objects''.
\newblock In Proceedings 2001 ICRA. IEEE International Conference on Robotics
  and Automation, Vol.~3, Ieee, pp.~2642--2648.

\bibitem{akin2002development}
Akin, D.~L., Carignan, C.~R., and Foster, A.~W., 2002.
\newblock ``Development of a four-fingered dexterous robot end effector for
  space operations''.
\newblock In Proceedings 2002 IEEE International Conference on Robotics and
  Automation, Vol.~3, IEEE, pp.~2302--2308.

\bibitem{crisman1996graspar}
Crisman, J.~D., Kanojia, C., and Zeid, I., 1996.
\newblock ``Graspar: A flexible, easily controllable robotic hand''.
\newblock {\em IEEE Robotics \& Automation Magazine, \textbf{ 3}}(2),
  pp.~32--38.

\bibitem{laliberte1998simulation}
Lalibert{\'e}, T., and Gosselin, C.~M., 1998.
\newblock ``Simulation and design of underactuated mechanical hands''.
\newblock {\em Mechanism and machine theory, \textbf{ 33}}(1-2), pp.~39--57.

\bibitem{cutkosky1989grasp}
Cutkosky, M.~R., et~al., 1989.
\newblock ``On grasp choice, grasp models, and the design of hands for
  manufacturing tasks''.
\newblock {\em IEEE Transactions on robotics and automation, \textbf{ 5}}(3),
  pp.~269--279.

\bibitem{mccarthy2010geometric}
McCarthy, J.~M., and Soh, G.~S., 2010.
\newblock {\em Geometric design of linkages}, Vol.~11.
\newblock Springer Science \& Business Media.

\bibitem{todhunter1863spherical}
Todhunter, I., 1863.
\newblock {\em Spherical trigonometry, for the use of colleges and schools:
  with numerous examples}.
\newblock Macmillan.

\bibitem{birglen2007underactuated}
Birglen, L., Lalibert{\'e}, T., and Gosselin, C.~M., 2007.
\newblock {\em Underactuated robotic hands}, Vol.~40.
\newblock Springer.

\bibitem{birglen2004kinetostatic}
Birglen, L., and Gosselin, C.~M., 2004.
\newblock ``Kinetostatic analysis of underactuated fingers''.
\newblock {\em IEEE Transactions on Robotics and Automation, \textbf{ 20}}(2),
  pp.~211--221.

\bibitem{mazzotti2017dimensional}
Mazzotti, C., Troncossi, M., and Parenti-Castelli, V., 2017.
\newblock ``Dimensional synthesis of the optimal rssr mechanism for a set of
  variable design parameters''.
\newblock {\em Meccanica, \textbf{ 52}}(10), pp.~2439--2447.

\end{thebibliography}
\end{document}